\documentclass[11pt,a4paper]{article}
\usepackage[hyperref]{acl2020}
\usepackage{times}
\usepackage{latexsym}

\usepackage{microtype}

\usepackage{amsmath, amssymb, amsthm}
\usepackage{bbm}
\usepackage{booktabs}
\usepackage{graphicx}
\usepackage{makecell}
\usepackage{hhline}

\usepackage{xcolor}

\usepackage{pgfplotstable}
\usepackage{pgfplots}

\usepackage{url}
\usepackage[normalem]{ulem}

\newcommand\sL{\ensuremath{\mathcal{L}}}

\newcommand\bv{\ensuremath{\mathbf{v}}}

\newcommand\bx{\ensuremath{\mathbf{x}}}
\newcommand\by{\ensuremath{\mathbf{y}}}

\newcommand\BE{\ensuremath{\mathbb{E}}}

\newcommand\T{\text}

\newcommand\refeqn[1]{(\ref{eqn:#1})}

\newcommand\refsec[1]{Section~\ref{sec:#1}}

\newcommand\reffig[1]{Figure~\ref{fig:#1}}

\newcommand\reftab[1]{Table~\ref{tab:#1}}
\newcommand\refapp[1]{Appendix~\ref{sec:#1}}

\usepackage{color}

\newcommand{\fone}{$\T{F}_1$}
\newcommand\ph[1]{\phantom{#1}}

\newcommand{\rlr}{$\text{RL}_{\text{R}}$}
\newcommand{\rlc}{$\text{RL}_{\text{C}}$}
\newcommand{\rlrc}{$\text{RL}_{\text{R+C}}$}

\definecolor{lblue}{HTML}{A6CEE3}
\definecolor{lgreen}{HTML}{B2DF8A}
\definecolor{lred}{HTML}{FB9A99}
\definecolor{lorange}{HTML}{FDBF6F}
\definecolor{mblue}{HTML}{80B1D3}
\definecolor{mgreen}{HTML}{B3DE69}
\definecolor{mred}{HTML}{FB8072}
\definecolor{morange}{HTML}{FDB462}
\definecolor{blue}{HTML}{1F78B4}
\definecolor{green}{HTML}{33A02C}
\definecolor{red}{HTML}{E31A1C}
\definecolor{orange}{HTML}{FF7F00}
\definecolor{dblue}{HTML}{08519C}
\definecolor{dgreen}{HTML}{006D2C}
\definecolor{dorange}{HTML}{EC7014}

\newcommand{\dblue}[1]{{\color{dblue} #1}}

\newcommand{\dorange}[1]{{\color{dorange} #1}}

\newcommand{\ul}[1]{\uline{#1}}

\aclfinalcopy % Uncomment this line for the final submission
\def\aclpaperid{1009} %  Enter the acl Paper ID here

\title{Optimizing the Factual Correctness of a Summary:\\
A Study of Summarizing Radiology Reports}

\author{Yuhao Zhang$^1$, Derek Merck$^2$, Emily Bao Tsai$^1$, \\
\bf{Christopher D. Manning$^1$, Curtis P. Langlotz$^1$} \\
  $^1$Stanford University\quad$^2$University of Florida\\
  \tt{ \{yuhaozhang, ebtsai, manning, langlotz\}@stanford.edu }\\
  \tt{ derek.merck@ufl.edu }\\}

\date{}

\begin{document}
\maketitle

%%%%%
% slightly compress the space between equations and text
% \setlength{\abovedisplayskip}{8pt}
% \setlength{\belowdisplayskip}{8pt}
%%%%%

\begin{abstract}
Neural abstractive summarization models are able to generate summaries which have high overlap with human references.
However, existing models are not optimized for factual correctness, a critical metric in real-world applications.
In this work, we develop a general framework where we evaluate the factual correctness of a generated summary by fact-checking it automatically against its reference using an information extraction module.
We further propose a training strategy which optimizes a neural summarization model with a factual correctness reward via reinforcement learning.
We apply the proposed method to the summarization of radiology reports, where factual correctness is a key requirement.
On two separate datasets collected from hospitals, we show via both automatic and human evaluation that the proposed approach substantially improves the factual correctness and overall quality of outputs over a competitive neural summarization system, producing radiology summaries that approach the quality of human-authored ones.
\end{abstract}
\section{Introduction}

\begin{figure}[t]
    \centering
    \small
\def\arraystretch{1.5}
\begin{tabular}{ | p{0.43\textwidth} | }
\hline
{\bf Background:}
radiographic examination of the chest. clinical history: 80 years of age, male ...
\vspace{0.5em}
\newline
{\bf Findings}:
frontal radiograph of the chest demonstrates repositioning of the right atrial lead possibly into the ivc.
...
a right apical pneumothorax can be seen from the image. moderate right and small left pleural effusions continue. no pulmonary edema is observed. heart size is upper limits of normal.
\\
\hhline{|=|}
{\bf Human Summary}: pneumothorax is seen. bilateral pleural effusions continue.
\\ 
\hhline{|=|}
{\bf Summary A} (ROUGE-L = 0.77):\newline
\dorange{no pneumothorax is observed}. bilateral pleural effusions continue.
\\ 
\hline
{\bf Summary B} (ROUGE-L = 0.44):\newline
\dblue{pneumothorax is observed} on radiograph. bilateral pleural effusions continue to be seen.
\\
\hline

\end{tabular}
    \caption{A (truncated) radiology report and summaries with their ROUGE-L scores. Compared to the human summary, Summary A has high textual overlap (i.e., ROUGE-L) but makes \dorange{a factual error}; Summary B has a lower ROUGE-L score but is \dblue{factually correct}.}
    \label{fig:intro-example}
\end{figure}

Neural abstractive summarization systems aim at generating sentences which compress a document while preserving the key facts in it~\cite{nallapati2016abstractive,see2017get,chen2018fast}.
These systems are potentially useful in many real-world applications.
For example, \citet{zhang2018radsum} have shown that customized neural abstractive summarization models are able to generate radiology summary statements with high quality by summarizing textual findings written by radiologists.
This task has significant clinical value because of its potential to accelerate the radiology workflow, reduce repetitive human labor, and improve clinical communications~\cite{kahn2009toward}.

However, while existing abstractive summarization models are optimized to generate summaries that highly overlap with human references~\cite{paulus2018a}, this does not guarantee factually correct summaries, as shown in~\reffig{intro-example}.
Therefore, maintaining factual correctness of the generated summaries remains a critical yet unsolved problem.
For example, \citet{zhang2018radsum} found that about 30\% of the outputs from a radiology summarization model contain factual errors or inconsistencies.
This has made such a system unusable in practice, as factual correctness is critically important in this domain to prevent medical errors.

Existing attempts at improving the factual correctness of abstractive summarization models have seen very limited success.
For example, \citet{cao2017faithful} augmented the attention mechanism of neural models with factual triples extracted with open information extraction systems;
\citet{falke2019ranking} studied using natural language inference systems to rerank generated summaries based on their factual consistencies;
\citet{kryciski2019evaluating} proposed to verify factual consistency of generated summaries with a weakly-supervised model.
Despite these efforts, none of the existing work has focused explicitly on optimizing an abstractive summarization system with a correctness objective.
As a result, even state-of-the-art systems trained with ample data still produce summaries with a substantial number of factual errors~\cite{goodrich2019assessing,kryciski2019neural}.

In this work we aim to optimize the factual correctness of existing neural summarization systems, with a focus on summarizing radiology reports.
This task has several key properties that make it ideal for studying factual correctness in summarization models.
First, the clinical facts or observations present in radiology reports have less ambiguity compared to open-domain text, which allows objective comparison of facts.
Second, radiology reports involve a relatively limited space of facts, which makes automatic measurement of factual correctness in the generated text approachable.
Lastly, as factual correctness is a crucial metric in this domain, improving factual correctness will directly lead to an ability to use the system.

To this end, we design a framework where an external information extraction system is used to extract information in the generated summary and produce a factual accuracy score by comparing it against the human reference summary.
We further develop a training strategy where we combine a factual correctness objective, a textual overlap objective and a language model objective, and jointly optimize them via reinforcement learning (RL).

On two datasets of radiology reports collected from different hospitals, we show that our training strategy substantially improves the factual correctness of the summaries generated by a competitive neural summarization system.
Moreover, we observe for the first time that, even in the absence of a factual correctness objective, optimizing a textual overlap-based metric substantially improves the factual correctness of the resulting system compared to maximum likelihood training.
We further show via human evaluation and analysis that our training strategy leads to summaries with higher overall quality and correctness and which are closer to the human-written ones.

Our main contributions are:
(i) we propose a general framework and a training strategy for improving the factual correctness of summarization models by optimizing a multi-part objective via RL;
(ii) we apply the proposed strategy to radiology reports, and empirically show that it improves the factual correctness of the generated summaries;
and (iii) we demonstrate via radiologist evaluation that our system is able to generate summaries with clinical validity close to human-written ones.
To our knowledge, our work represents the first attempt at directly optimizing a neural summarization system with a factual correctness objective via RL.
\section{Related Work}

\paragraph*{Neural Summarization Systems.}
Neural models for text summarization can be broadly divided into extractive approaches~\cite{cheng2016neural,nallapati2016summarunner} and abstractive approaches~\cite{nallapati2016abstractive,see2017get}.
While existing models are often trained in an end-to-end manner by maximizing the likelihood of the reference summaries, RL has been shown useful in recent work~\cite{chen2018fast, dong2018banditsum}.
Specifically, \citet{paulus2018a} found that directly optimizing an abstractive summarization model on the ROUGE metric via RL can improve the summary ROUGE scores.
Our work extends the rewards used in existing work with a factual correctness reward to further improve the correctness of the generated summaries.

\paragraph*{Factual Correctness in Summarization.}
Our work is closely related to recent work that studies factual correctness in summarization.
\citet{cao2017faithful} proposed to improve summarization models by attending to fact triples extracted using open information extraction systems.
\citet{goodrich2019assessing} compared different information extraction systems to evaluate the factual accuracy of generated text.
\citet{falke2019ranking} explored using natural language inference systems to evaluate the correctness of generated summaries, and found models trained on existing datasets to be inadequate.
\citet{kryciski2019evaluating} proposed to evaluate factual consistencies in the generated summaries using a weakly-supervised fact verification model.
Despite these efforts, none of this work has shown success in directly optimizing a summarization system for factual correctness, and to our knowledge our work represents the first attempt in this direction.
While our work is focused on improving neural summarization models, we note that the idea of using information extraction systems to evaluate the fidelity of generated text has also been explored for data-to-text generation~\cite{wiseman2017challenges,dhingra2019handling}.

\paragraph*{Summarization of Radiology Reports.}
\citet{zhang2018radsum} first studied the problem of automatic generation of radiology impressions by summarizing textual radiology findings, and showed that an augmented pointer-generator model achieves high overlap with human references.
\citet{macavaney2019ontology} extended this model with an ontology-aware pointer-generator and showed improved summarization quality.
\citet{li2019hybrid} and \citet{liu2019clinically} studied generating textual descriptions of radiology findings from medical images, and proposed RL-based approaches to tackle this problem.
While \citet{zhang2018radsum} found that about 30\% of the radiology summaries generated from neural models contain factual errors, improving factual correctness in radiology summarization remains unstudied.

\section{Task \& Baseline Pointer-Generator}
\label{sec:baseline}

We start by briefly introducing the task of summarizing radiology findings. 
Given a passage of radiology findings represented as a sequence of tokens $\bx = \{ x_1, x_2, \ldots, x_N\}$, with $N$ being the length of the findings, the task involves finding a sequence of tokens $\by = \{y_1, y_2, \ldots, y_L\}$ that best summarizes the salient and clinically significant findings in $\bx$.
In routine radiology workflow, an output sequence $\by$ is produced by the radiologist, which we treat as a reference summary sequence.%
\footnote{While the name ``impression'' is often used in clinical settings, we use ``summary'' and ``impression'' interchangeably.}

To model the summarization process, we use the background-augmented pointer-generator network~\cite{zhang2018radsum} as the backbone of our method. 
This abstractive summarization model extends a pointer-generator~\cite{see2017get} with a separate background section encoder and is shown to be effective in summarizing radiology notes with multiple sections. 
We briefly describe this model and refer readers to the original papers for details.

At a high level, this model first encodes the input sequence $\bx$ into hidden states with a Bi-directional Long Short-Term Memory (Bi-LSTM) network, and then generates an output sequence $\by$ with a separate LSTM decoder.
To make the input information available at decoding time, an attention mechanism~\cite{bahdanau2014neural} over the input hidden states is also added to the decoder.

The baseline pointer-generator model by~\citet{zhang2018radsum} adds two augmentations to this attentional encoder-decoder model to make it suitable for summarizing radiology findings:

\paragraph*{Copy Mechanism.} 
To enable the model to copy words from the input, a copy mechanism~\cite{vinyals2015pointer,see2017get} is added to calculate a generation probability at each step of decoding.
This generation probability is then used to blend the original output vocabulary distribution and a copy distribution to generate the next word.

\paragraph*{Background-guided Decoding.}
As shown in \reffig{intro-example}, radiology reports often consist of a background section which documents the crucial study background information (e.g., purpose of the study, patient conditions), and a findings section which documents clinical observations.
While words can be copied from the findings section to form the summary, \citet{zhang2018radsum} found it worked better to separately encode the background section, and inject the representation into the decoding process by concatenating it with the input.

\section{Fact Checking in Summarization}
\label{sec:fact-checking}

Summarization models such as the one described in \refsec{baseline} are commonly trained with the teacher-forcing algorithm~\cite{williams1989learning} by maximizing the likelihood of the reference, human-written summaries.
However, this training strategy results in a significant discrepancy between what the model sees during training and test time, often referred to as the \textit{exposure bias} issue~\cite{ranzato2015sequence}, leading to degenerate output at test time.

\begin{figure*}[t]
  \centering
  \includegraphics[width=\textwidth]{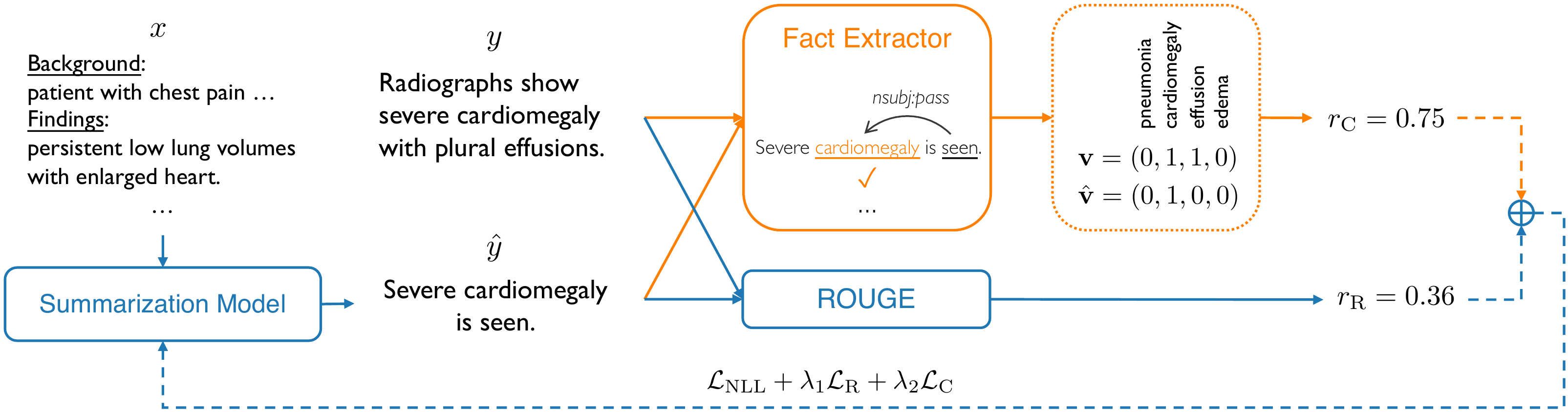}
  \caption{Our proposed training strategy. Compared to existing work which relies only on a ROUGE reward $r_\text{R}$, we add a factual correctness reward $r_\text{C}$ which is enabled by a fact extractor. The summarization model is updated via RL, using a combination of the NLL loss, a ROUGE-based loss and a factual correctness-based loss. For simplicity we only show a subset of the clinical variables in the fact vectors $\bv$ and $\hat{\bv}$.}
  \label{fig:diagram}
\end{figure*}

An alternative training strategy is to directly optimize standard metrics such as ROUGE scores~\cite{lin2004rouge} with RL and this was shown to improve summarization quality~\cite{paulus2018a}.
Nevertheless, this method still provides no guarantee that the generated summary is factually accurate and complete, since the ROUGE scores merely measure the superficial text overlap between two sequences and do not account for the factual alignment between them.
To illustrate this, a reference sentence \emph{pneumonia is seen} and a generated sentence \emph{pneumonia is \ul{not} seen} have substantial text overlap and thus the generated sentence would achieve a high ROUGE score, however the generated sentence conveys an entirely opposite fact.
In this section we first introduce a method to verify the factual correctness of the generated summary against the reference summary, and then describe a training strategy to directly optimize a factual correctness objective to improve summary quality.

\subsection{Evaluating Factual Correctness via Fact Extraction}
A convenient way to explicitly measure the factual correctness of a generated summary against the reference is to first extract and represent the facts in a structured format.
To this end, we define a \textit{fact extractor} to be an information extraction (IE) module, denoted as $f$, which takes in a summary sequence $y$ and returns a structured fact vector $\bv$:
\begin{equation}
  \bv = f(y) = (v_1, ..., v_m)
\end{equation}
where $v_i$ is a categorical variable that we want to measure via fact checking and $m$ the total number of such variables. 
For example, in the case of summarizing radiology reports, $v_i$ can be a binary variable that describes whether an event or a disease such as \emph{pneumonia} is present or not in a radiology study.

Given a fact vector $\bv$ output by $f$ from a reference summary and $\hat{\bv}$ from a generated summary, we further define a \textit{factual accuracy} score $s$ to be the ratio of variables in $\hat{\bv}$ which equal the corresponding variables in $\bv$, namely:
\begin{equation}
  s(\hat{\bv}, \bv) = \frac{\sum_{i=1}^m \mathbbm{1} [v_i = \hat{v}_i] }{m}
\label{eqn:factual-accuracy}
\end{equation}
where $s \in [0,1]$.
Note that this method requires a summary to be both precise and complete in order to achieve a high $s$ score:
missing out a positive variable or falsely claiming a negative variable will be equally penalized.

Our general definition of the fact extractor module $f$ allows it to have different realizations for different domains. For our task of summarizing radiology findings, we make use of the open-source CheXpert radiology report labeler~\cite{irvin2019chexpert}.%
\footnote{\url{https://github.com/stanfordmlgroup/chexpert-labeler}}
At its core, the CheXpert labeler parses the input sentences into dependency structures and runs a series of surface and syntactic rules to extract the presence status of 14 clinical observations seen in chest radiology reports.%
\footnote{For this study we used a subset of these variables and discuss the reasons in \refapp{variables}.}
It was evaluated to have over 95\% overall \fone{} when compared against oracle annotations from multiple radiologists on a large-scale radiology report dataset.

\subsection{Improving Factual Correctness via Policy Learning}

The fact extractor module introduced above not only enables us to measure the factual accuracy of a generated summary, but also provides us with an opportunity to directly optimize the factual accuracy as an objective.
This can be achieved by viewing our summarization model as an agent, the actions of which are to generate a sequence of words to form the summary $\hat{y}$, conditioned on the input $x$.%
\footnote{For clarity, we drop the bold symbol and use $x$ and $y$ to represent the input and output sequences, respectively.}
The agent then receives rewards $r(\hat{y})$ for its actions, where the rewards can be designed to measure the quality of the generated summary.
Our goal is to learn an optimal policy $P_\theta(y | x)$ for the summarization model, parameterized by the network parameters $\theta$, which achieves the highest expected reward under the training data.

Formally, we minimize loss $\sL$, the negative expectation of the reward $r(\hat{y})$ over the training data:
\begin{equation}
\sL(\theta) = - \BE_{\hat{y} \sim P_{\theta}(y | x)} [r(\hat{y})].
\end{equation}
The gradient can be calculated as \cite[REINFORCE][]{williams1992simple}:
\begin{equation}
\nabla_\theta \mathcal{L}(\theta) = - \mathbb{E}_{\hat{y} \sim P_{\theta}(y | x)} [\nabla_\theta \log{P_{\theta}(\hat{y} | x)} r(\hat{y})].
\label{eqn:gradient}
\end{equation}
In practice, we approximate this gradient over a training example with a single Monte Carlo sample and deduct a baseline reward to reduce the variance of the gradient estimation:
\begin{equation}
\nabla_\theta \mathcal{L}(\theta) \approx - \nabla_\theta \log{P_{\theta}(\hat{y}_s | x)} ( r(\hat{y}_s) - \bar{r}),
\label{eqn:self-critical}
\end{equation}
where $\hat{y}_s$ is a sampled sequence from the model and $\bar{r}$ a baseline reward. 
Here we adopt the \emph{self-critical training} strategy~\cite{rennie2017self}, where we obtain the baseline reward $\bar{r}$ by applying the same reward function $r$ to a greedily decoded sequence $\hat{y}_g$, i.e., $\bar{r} = r(\hat{y}_g)$.
We empirically find that using this self-critical baseline reward helps stabilize the training of our summarization model.

\subsection{Reward Function}

The learning strategy in Equation~\refeqn{self-critical} provides us with the flexibility to optimize arbitrary reward functions.
Here we decompose our reward function into two parts:
\begin{equation}
r = \lambda_1 r_\text{R} + \lambda_2 r_\text{C},
\end{equation}
where $r_\text{R} \in [0,1]$ is a ROUGE reward, namely the ROUGE-L score~\cite{lin2004rouge} of the predicted sequence $\hat{y}$ against the reference $y$;
$r_\text{C} \in [0,1]$ is a correctness reward, namely the factual accuracy $s$ of the predicted sequence against the reference sequence, as in Equation~\refeqn{factual-accuracy};
$\lambda_1, \lambda_2 \in [0,1]$ are scalar weights that control the balance between the two.
To measure the similarity between the reference and the generation, we also experimented with more recent metrics that rely on neural representations of text, such as the BERTScore \cite{zhang2019bertscore}. However, we found that these metrics, mostly trained on web and newswire data, generalize poorly to our domain of text.

\citet{paulus2018a} found that directly optimizing a reward function without the original negative log-likelihood (NLL) objective as used in teacher-forcing can hurt the readability of the generated summaries, and proposed to alleviate this problem by combining the NLL objective with the RL loss.
Here we adopt the same strategy, and our final loss during training is:
\begin{equation}
  \sL = \lambda_1 \sL_\text{R} + \lambda_2 \sL_\text{C} + \lambda_3 \sL_\text{NLL},
  \label{eqn:final-loss}
\end{equation}
where $\lambda_3 \in [0,1]$ is an additional scalar that controls the weight of the NLL loss.

Our overall training strategy is illustrated in \reffig{diagram}. 
Our final loss jointly optimizes three aspects of the summaries: 
$\sL_\text{NLL}$ serves as a conditional language model that optimizes the fluency and relevance of the generated summary, $\sL_\text{R}$ controls the brevity of the summary and encourages summaries which have high overlap with human references, and $\sL_\text{C}$ encourages summaries that are factually accurate when compared against human references.

\section{Experiments}

We collected two real-world radiology report datasets and describe our experiments using them as our main training and evaluation corpora.

\subsection{Data Collection}

We collected anonymized chest radiographic reports within a certain period of time from two collaborating hospitals: the Stanford University Hospital and the Rhode Island Hospital (RIH).%
\footnote{Our retrospective study has been approved by the corresponding institutional review boards with waiver of consent.}

For both datasets, we ran simple preprocessing following \citet{zhang2018radsum}.
To test the generalizability of the models, instead of using random stratification, we stratified each dataset over time into training, dev and test splits. We include statistics of both datasets in \reftab{dataset-stats} and preprocessing and stratification details in \refapp{data-preprocess}.

\begin{table}[t]
% \small
    \centering
    \begin{tabular}{lcc}
      \toprule
      & \multicolumn{2}{c}{Number of Examples} \\
      Split & Stanford & RIH \\
      \midrule
      Train & 89,992 (68.8\%) & 84,194 (60.3\%) \\
      Dev & 22,031 (16.8\%) & 25,966 (18.6\%) \\
      Test & 18,827 (14.4\%) & 29,494 (21.1\%) \\
      \midrule
      Total & 130,850 & 139,654 \\
      \bottomrule
    \end{tabular}
    \caption{Statistics of the Stanford and RIH datasets.}
  \label{tab:dataset-stats}
\end{table}

% actual results
\begin{table*}[t]
% \small
  \centering
  \setlength{\tabcolsep}{0.4em}
  \begin{tabular}{l@{\hskip 1.5em}cccc@{\hskip 1.5em}cccc}
    \toprule
    & \multicolumn{4}{c}{\textbf{Stanford}} & \multicolumn{4}{c}{\textbf{RIH}}\\
    System   & R-1 & R-2 & R-L & Factual \fone & R-1 & R-2 & R-L & Factual \fone \\
    \midrule
    LexRank~\citep{erkan2004lexrank} & 26.8 & 16.3 & 23.6 & --- & 20.6 & 10.7 & 18.3 & --- \\
    BanditSum~\citep{dong2018banditsum} & 32.7 & 20.9 & 29.0 & --- & 26.1 & 14.0 & 23.3 & --- \\
    \midrule
    PG Baseline~\citep{zhang2018radsum} & 48.3 & 38.8 & 46.6 & 55.9 & 54.1 & 44.7 & 52.2 & 69.3 \\
    \midrule
    PG + \rlr & \bf{52.0} & \bf{41.1} & \bf{49.5} & 63.2 & \bf{58.0} & \bf{47.2} & \bf{55.7} & 73.3 \\
    PG + \rlc & 50.7 & 39.7 & 48.0 & \bf{65.9} & 55.2 & 45.4 & 52.9 & \bf{75.4} \\
    PG + \rlrc & \bf{52.0} & 41.0 & 49.3 & 64.5 & 57.0 & 46.6 & 54.7 & 74.8 \\
    \bottomrule
  \end{tabular}
  \caption{Main results on the two datasets.
  R-1, R-2, R-L represent the ROUGE scores. 
  PG Baseline represents our baseline augmented pointer-generator; \rlr{}, \rlc{} and \rlrc{} represent RL training with the ROUGE reward alone, with the factual correctness reward alone and with both. 
  All the ROUGE scores have a 95\% confidence interval of at most $\pm$0.6. 
  \fone{} scores for extractive models were not evaluated for the reason discussed in \refsec{eval}.}
\label{tab:main-results}
\end{table*}

\subsection{Models}

As we use the augmented pointer-generator network described in~\refsec{baseline} as the backbone of our method, we mainly compare against it as the baseline model (PG Baseline), and use the open implementation by~\citet{zhang2018radsum}.

For the proposed RL-based training, we compare three variants: training with only the ROUGE reward (\rlr), with only the factual correctness reward (\rlc), or with both (\rlrc).
All three variants have the NLL component in the training loss as in Equation~\refeqn{final-loss}.
For all variants, we initialize the model with the best baseline model trained with standard teacher-forcing, and then finetune it on the training data with the corresponding RL loss, until it reaches the best validation score.

To understand the difficulty of the task and evaluate the necessity of using abstractive summarization models, we additionally evaluate two extractive summarization methods:
(1) LexRank~\cite{erkan2004lexrank}, a widely-used non-neural extractive summarization algorithm; and
(2) BanditSum~\cite{dong2018banditsum}, a state-of-the-art RL-based neural extractive summarization model.
For both methods we use their open implementations.
We include other model implementation and training details in \refapp{implementation}.

\subsection{Evaluation}
\label{sec:eval}

We use two sets of metrics to evaluate model performance at the corpus level. 
First, we use the standard \textbf{ROUGE} scores~\cite{lin2004rouge}, and report the \fone{} scores for ROUGE-1, ROUGE-2 and ROUGE-L, which compare the word-level unigram, bigram and longest common sequence overlap with the reference summary, respectively.

For factual correctness evaluation, we use a \textbf{Factual \fone{}} score.
While the factual accuracy score $s$ that we use in the reward function evaluates how factually accurate a specific summary is, comparing it at the corpus level can be misleading, for the same reason that accuracy is a misleading measure in information retrieval \cite{manning2008introduction}.
To understand this, imagine the case where a clinical variable $v$ has rare presence in the corpus.
A model which always generates a negative summary for it (i.e., $v = 0$; the disease is not present) can have high accuracy, but is useless in practice.
Instead, for each variable, we obtain a model's predictions over all test examples and calculate its \fone{} score.
We then macro-average the \fone{} of all variables to obtain the overall factual \fone{} score of the model.

Note that the CheXpert labeler that we use is specifically designed to run on radiology summaries, which usually have a different style of language compared to the radiology findings section of the reports (see further analysis in \refsec{analysis}). 
As a result, we found the labeler to be less accurate when applied to the findings section.
For this reason, we were not able to estimate the factual \fone{} scores on the summaries generated by the two extractive summarization models.

\section{Results}
\label{sec:results}

We first present our automatic evaluation results on the two collected datasets. We then present a human evaluation with board-certified radiologists where we compare the summaries generated by humans, the baseline and our proposed model.

\subsection{Automatic Evaluation}

Our main results on both datasets are shown in \reftab{main-results}.
We first notice that while the neural extractive model, BanditSum, outperforms the non-neural extractive method on ROUGE scores, our PG baseline model substantially outperforms both of them, suggesting that on both datasets abstractive summarization is necessary to generate summaries comparable to human-written ones.
We further show that this difference is likely due to the different styles of language (see \refsec{analysis}): while radiologists tend to use more compressed language when writing the summaries, extractive methods produce more verbose summaries that fail to capture this difference.

% fine-grained results
\begin{table}[t]
  % \small
    \centering
    \setlength{\tabcolsep}{0.2em}
    \begin{tabular}{lccc}
      \toprule
      Variable & PG Baseline & \rlrc & $\Delta$ \\
      \midrule
      No Finding & 77.3 & 81.5 & \ph{0}+4.2$^*$ \\
      Cardiomegaly & 29.5 & 40.4 & +10.9$^*$ \\
      Airspace Opacity & 64.6 & 74.9 & +10.3$^*$ \\
      Edema & 58.4 & 70.9 & +12.5$^*$ \\
      Consolidation & 46.3 & 53.2 & \ph{0}+6.9$^*$ \\
      Pneumonia & 46.7 & 46.8 & \ph{0}+0.2\ph{$^*$} \\
      Atelectasis & 48.8 & 56.3 & \ph{0}+7.5$^*$ \\
      Pneumothorax & 69.5 & 82.9 & +13.4$^*$ \\
      Pleural Effusion & 62.0 & 73.4 & +11.4$^*$ \\
      \midrule
      Macro Avg. & 55.9 & 64.5 & \ph{0}+8.6$^*$ \\
      \bottomrule
    \end{tabular}
    \caption{Test set factual \fone{} scores for each variable on the Stanford dataset. $*$ marks statistically significant improvements with $p < .01$ under a bootstrap test.}
  \label{tab:var-stanford}
  \end{table}

On the Stanford dataset, training the pointer-generator model with ROUGE reward alone (\rlr) leads to improvements on all ROUGE scores, with a gain of 2.9 ROUGE-L scores.
Training with the factual correctness reward alone (\rlc) leads to the best overall factual \fone{} with a substantial gain of 10\% absolute, however with consistent decline in the ROUGE scores compared to \rlr{} training.
Combining the ROUGE and the factual correctness rewards (\rlrc) achieves a balance between the two, leading to an overall improvement of 2.7 on ROUGE-L and 8.6\% on factual \fone{} compared to the baseline.
This indicates that \rlrc{} training leads to both higher overlap with references and improved factual correctness.

Most surprisingly, while ROUGE has been criticized for its poor correlation with human judgment of quality and insufficiency for evaluating correctness of the generated text~\cite{chaganty2018price}, we find that optimizing ROUGE reward jointly with NLL leads to substantially more factually correct summaries than the baseline,
shown by the notable gain of 7.3\% factual \fone{} from the \rlr{} training.

All of our findings are consistent on the RIH dataset, with \rlrc{} achieving an overall improvement of 2.5 ROUGE-L and 5.5\% factual \fone{} scores.

\paragraph*{Fine-grained Correctness.}
To understand how improvements in individual variables contribute to the overall improvement, we show the fine-grained factual \fone{} scores for all variables on the Stanford dataset in~\reftab{var-stanford} and include results on the RIH dataset in \refapp{var-rih}.
We find that on both datasets, improvements in \rlrc{} can be observed on all variables tested.
We further find that, as we change the initialization across different training runs, while the overall improvement on factual \fone{} stays approximately unchanged, the distribution of the improvement on different variables can vary substantially.
Developing a training strategy for fine-grained control over different variables is an interesting direction for future work.

% qualitative examples
\begin{figure}[t]
  \centering
	% \small
\scriptsize
\def\arraystretch{1.5}
\begin{tabular}{ | p{0.45\textwidth} | }
\hline
Stanford Dataset\\
\hline
% test example 16345
{\bf Background:}
radiographic examination of the chest ...
\vspace{0.5em}
\newline
{\bf Findings}:
continuous rhythm monitoring device again seen projecting over the left heart. persistent low lung volumes with unchanged cardiomegaly. again seen is a diffuse reticular pattern with interstitial prominence demonstrated represent underlying emphysematous changes with superimposed increasing moderate pulmonary edema. small bilateral pleural effusions. persistent bibasilar opacities left greater than right which may represent infection versus atelectasis.
\\
\hline
{\bf Human}: increased moderate \dblue{\ul{pulmonary edema}} with small bilateral \dblue{\ul{pleural effusions}}. left greater than right \dblue{\ul{basilar opacities}} which may represent \dblue{\ul{infection}} versus atelectasis.
\\
\hline
{\bf PG Baseline} ($s=0.33$): no significant interval change.
\\
\hline
{\bf \rlrc{}} ($s = 1.00$): increasing moderate \dblue{\ul{pulmonary edema}}. small bilateral \dblue{\ul{pleural effusions}}. persistent \dblue{\ul{bibasilar opacities}} left greater than right which may represent \dblue{\ul{infection}} versus atelectasis.
\\ 

\hhline{|=|}

RIH Dataset
\\
\hline
{\bf Background:}
history: lobar pneumonia, unspecified organism ...
\vspace{0.5em}
\newline
{\bf Findings}:
lines/tubes: none. lungs: \dorange{\uwave{right middle lobe airspace disease}} seen on prior radiographs from $<$date$>$ and $<$date$>$ is \dorange{\uwave{no longer evident}}. bilateral lungs appear clear. pleura: there is no pleural effusion or pneumothorax. heart and mediastinum: no cardiomegaly. thoracic aorta appears calcified and mildly tortuous. 
bones: ...
\\
\hline
{\bf Human}: no acute cardiopulmonary abnormality.
\\
\hline
{\bf PG Baseline} ($s=0.75$): \dorange{\uwave{right middle lobe airspace disease}} could represent atelectasis, aspiration or pneumonia.
\\
\hline
{\bf \rlrc{}} ($s = 1.00$): no acute cardiopulmonary abnormality.
\\ 
\hline

\end{tabular}
  \caption{Truncated examples from the test sets along with human, PG baseline and \rlrc{} outputs. 
  Factual accuracy scores ($s$) are also shown for the model outputs. 
  For the Stanford example, \dblue{\ul{clinical observations}} in the summaries are marked for clarity; 
  for RIH, \dorange{\uwave{a wrongly copied observation}} is marked.}
  \label{fig:main-examples}
\end{figure}

\paragraph*{Qualitative Results.}
In \reffig{main-examples} we present two example reports along with the human references, the PG baseline outputs and \rlrc{} outputs.
In the first example, while baseline output seems generic and does not include any meaningful observation, the summary from the \rlrc{} model aligns well with the reference, and therefore achieves a higher factual accuracy score.
In the second example, the baseline model wrongly copied an observation from the findings although the actual context is \emph{no longer evident}, while the \rlrc{} model correctly recognizes this and produces a better summary.

% human eval results
\begin{table}[t]
% \small
  \centering
  \setlength{\tabcolsep}{0.75em}
  \begin{tabular}{lccc}
    \toprule
    Metric & Win & Tie & Lose \\
    \midrule
    \multicolumn{4}{c}{Our Model vs. PG Baseline} \\
    \midrule
    Fluency & \phantom{0}7\% & 60\% & 33\% \\
    Factual Correctness & 31\% & 55\% & 14\% \\
    Overall Quality & 48\% & 24\% & 28\% \\
    \midrule
    \multicolumn{4}{c}{Our Model vs. Human Reference} \\
    \midrule
    Fluency & 17\% & 54\% & 29\% \\
    Factual Correctness & 23\% & 49\% & 28\% \\
    Overall Quality & 44\% & 17\% &  39\% \\
    \bottomrule
  \end{tabular}
  \caption{Results of the radiologist evaluation. The top three rows present results when comparing our \rlrc{} model output versus the baseline model output; the bottom three rows present results when comparing our model output versus the human-written summaries.}
\label{tab:human}
\end{table}

\subsection{Human Evaluation}
\label{sec:human-eval}

To study whether the improvements in the factual correctness scores lead to improvement in summarization quality under expert judgment, we run a comparative human evaluation following previous work~\cite{chen2018fast, dong2018banditsum,zhang2018radsum}.
We sampled 50 test examples from the Stanford dataset, and for each example we presented to two board-certified radiologists the full radiology findings along with blinded summaries from (1) the human reference, (2) the PG baseline and (3) our \rlrc{} model.
We shuffled the three summaries such that the correspondence cannot be guessed, and asked the radiologists to compare them based on the following three metrics: (1) \textbf{fluency}, (2) \textbf{factual correctness and completeness}, and (3) \textbf{overall quality}.
For each metric we asked the radiologists to rank the three summaries, with ties allowed.
After the evaluation, we converted each ranking into two binary comparisons: (1) our model versus the baseline model, and (2) our model versus human reference.

The results are shown in~\reftab{human}.
Comparing our model against the baseline model, we find that:
(1) in terms of fluency our model is less preferred, although a majority of the results (60\%) are ties;
(2) our model wins more on factual correctness and overall quality.
Comparing our model against human references, we find that:
(1) human wins more on fluency;
(2) factual correctness results are close, with 72\% of our model outputs being at least as good as human;
(3) surprisingly, in terms of overall quality our model was slightly preferred by the radiologists compared to human references.
Lastly, when comparing the baseline model against human references, we find that outputs from the baseline model are much less correct and lower-quality than human summaries.
\section{Analysis \& Discussion}
\label{sec:analysis}

% LM eval results
\begin{table}[t]
  % \small
    \centering
    \setlength{\tabcolsep}{0.8em}
    \begin{tabular}{lcc}
      \toprule
      System &  Stanford pplx. & RIH pplx.\\
      \midrule
      Human & \ph{0}6.7 & \ph{0}5.5 \\
      % Human + noise & \ph{0}8.7 & \ph{0}8.1 \\
      \midrule
      LexRank & 10.8 & 36.9 \\
      BanditSum & \ph{0}9.9 & 40.9 \\
      \midrule
      PG Baseline & \ph{0}4.8 & \ph{0}3.8 \\
      PG + \rlrc & \ph{0}6.5 & \ph{0}4.8 \\
      \bottomrule
    \end{tabular}
    \caption{Perplexity scores as evaluated by the trained radiology impression LM on the test set human references and model predictions.}
  \label{tab:language-model}
  \end{table}

\paragraph*{Fluency and Style of Summaries.}
Our human evaluation results in~\refsec{human-eval} suggest that in terms of fluency our model output is less preferred than human reference and baseline output.
To further understand the fluency and style of summaries from different models at a larger scale, we trained a neural language model (LM) for radiology summaries following previous work~\cite{liu2018generating}.
Intuitively, radiology summaries which are more fluent and consistent with humans in style should be able to achieve a lower perplexity under this in-domain LM, and vice versa.
To this end, we collected all human-written summaries from the training and dev split of both datasets, which in total gives us about 222,000 summaries.
We then trained a strong Mixture of Softmaxes LM~\cite{yang2017breaking} on this corpus, and evaluated the perplexity of test set outputs for all models.

The results are shown in \reftab{language-model}.
We find that while extractive models can achieve non-trivial overlap with references, their perplexity scores tend to be much higher than humans.
We conjecture that this is because radiologists are trained to write the summaries with more compressed language than when they are writing the findings, therefore sentences directly extracted from the findings tend to be more verbose than needed.

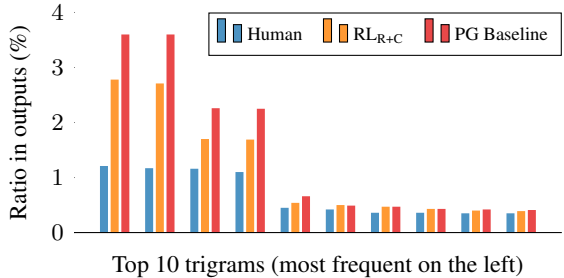
\begin{figure}[t]
  \begin{tikzpicture}
  \centering
  \begin{axis}[
        ybar=1pt,
        axis on top,
        height=4.5cm,
        width=8cm,
        bar width=3pt,
        ymajorgrids, tick align=inside,
        major grid style={draw=none},
        ymin=0, ymax=4.0,
        axis x line*=bottom,
        axis y line*=left,
        y axis line style={opacity=0},
        tickwidth=0pt,
        tick label style={font=\small},
        enlarge x limits=true,
        legend style={
            at={(1,1)},
            anchor=north east,
            legend columns=-1,
            font=\scriptsize,
            /tikz/every even column/.append style={column sep=0.2cm}
        },
        ylabel={Ratio in outputs (\%)},
        ylabel style={yshift=-15pt},
        xlabel={Top 10 trigrams (most frequent on the left)},
        xlabel style={yshift=10pt},
        label style={font=\small},
        symbolic x coords={
            1, 2, 3, 4, 5, 6, 7, 8, 9, 10},
        xtick=\empty,
        % xtick=data,
    %   nodes near coords={
    %     \pgfmathprintnumber[precision=3]{\pgfplotspointmeta}
    %   },
    %   every node near coord/.append style={font=\small}
    ]
    \addplot [draw=none, fill=blue!80] coordinates {
      (1, 1.21)
      (2, 1.17)
      (3, 1.16)
      (4, 1.10)
      (5, 0.45)
      (6, 0.42)
      (7, 0.36)
      (8, 0.36)
      (9, 0.35)
      (10, 0.35)};
    
    \addplot [draw=none, fill=orange!80] coordinates {
      (1, 2.78)
      (2, 2.71)
      (3, 1.70)
      (4, 1.69)
      (5, 0.54)
      (6, 0.50)
      (7, 0.47)
      (8, 0.43)
      (9, 0.40)
      (10, 0.39)};
    
    \addplot [draw=none, fill=red!80] coordinates {
      (1, 3.60)
      (2, 3.60)
      (3, 2.26)
      (4, 2.25)
      (5, 0.66)
      (6, 0.49)
      (7, 0.47)
      (8, 0.43)
      (9, 0.42)
      (10, 0.41)};

    \legend{Human, \rlrc, PG Baseline}
  \end{axis}
\end{tikzpicture}
  \caption{Distributions of the top 10 most frequent trigrams from model outputs on the Stanford test set.}
  \label{fig:ngram}
\end{figure}

We further observe that the baseline model achieves even lower perplexity than humans, and our proposed method leads to a perplexity score much closer to human references.
We hypothesize that this is because models trained with teacher-forcing are prone to generic generations which are fluent and relevant but may not be factually correct.
Training with the proposed rewards alleviates this issue, leading to summaries more consistent with humans in style.
For example, we find that \textit{no significant interval change} is a very frequent generation from the baseline, regardless of the actual input. 
This sentence occurs in 34\% of the baseline outputs on the Stanford dev set, while the number for \rlrc{} and human are only 24\% and 17\%.
This hypothesis is further confirmed when we plot the distribution of the top 10 most frequent trigrams from different models in~\reffig{ngram}:
while the baseline heavily reuses the few most frequent trigrams, our model \rlrc{} tends to have more diverse summaries which are closer to human references.
The same trends are observed for 4-grams and 5-grams.

\paragraph*{Limitations.}
While we showed the success of our proposed method on improving the factual correctness of a radiology summarization model, we also recognize several limitations of our work.
First, our proposed training strategy crucially depends on the availability of an external IE module.
While this IE module is relatively easy to implement for a domain with a limited space of facts, how to generalize this method to open-domain summarization remains unsolved.
Second, our study was based on a rule-based IE system, and the use of a more robust statistical IE model can potentially improve the results.
Third, we mainly focus on key factual errors which result in a flip of the binary outcome of an event (e.g., presence of disease), whereas factual errors in generated summaries can occur in other forms such as wrong adjectives or coreference errors~\cite{kryciski2019neural}.
We leave the study of these problems to future work.

\section{Conclusion}

In this work we presented a general framework and a training strategy to improve the factual correctness of neural abstractive summarization models.
We applied this approach to the summarization of radiology reports, and showed its success via both automatic and human evaluation on two separate datasets collected from hospitals.

Our general takeaways include:
(1) in a domain with a limited space of facts such as radiology reports, a carefully implemented IE system can be used to improve the factual correctness of neural summarization models via RL;
(2) even in the absence of a reliable IE system, optimizing the ROUGE metrics via RL can substantially improve the factual correctness of the generated summaries.

We hope that our work draws the community's attention to the factual correctness issue of abstractive summarization models and inspires future work in this direction.

\section*{Acknowledgments}

The authors would like to thank the anonymous reviewers, Peng Qi and Urvashi Khandelwal for their helpful comments, and Dr. Jonathan Movson for his help with obtaining the RIH data used in this study.

\bibliography{main}
\bibliographystyle{acl_natbib}

\clearpage
\appendix
\section{Clinical Variables Inclusion Criteria}
\label{sec:variables}

While the CheXpert labeler that we use is able to extract status for 14 clinical variables, we found that several variables are very rarely represented in our corpora and therefore using all of them makes the calculation of the factual \fone{} score very unstable. 
For example, we found that training the same model using different random initializations would result in highly varying \fone{} scores for these variables.
For this reason, for both datasets we removed from the factual \fone{} calculation all variables which have less than 3\% positive occurrences on the validation set.
We further removed the variables ``Pleural Other'' and ``Support Devices'' due to their ambiguity.
This process results in a total of 9 variables for the Stanford dataset and 8 for the RIH dataset.

Additionally, apart from the positive and negative status, the CheXpert labeler is also able to generate an \emph{uncertain} status for a variable, capturing observations with uncertainty, such as in the sentence ``\emph{pneumonia is likely represented}''. While we can modify the factual accuracy score to take uncertainty into account, for simplicity in this work we do not make the distinction between a positive status and an uncertain status.

\begin{table}[t]
  \centering
  \begin{tabular}{lcc}
    \toprule
    & \multicolumn{2}{c}{Time Coverage} \\
    Split & Stanford & RIH \\
    \midrule
    Train & 2009/01 -- 2014/04 & 2017/11 -- 2018/06 \\
    Dev & 2014/05 -- 2014/08 & 2018/07 -- 2018/09 \\
    Test & 2014/09 -- 2014/12 & 2018/10 -- 2018/12 \\
    \bottomrule
  \end{tabular}
  \caption{Time coverage of different splits in the Stanford and RIH datasets.}
\label{tab:time-coverage}
\end{table}

\section{Dataset Preprocessing and Stratification Details}
\label{sec:data-preprocess}

We preprocessed both the Stanford and the RIH datasets following \citet{zhang2018radsum}.
All reports were first tokenized with Stanford CoreNLP \cite{manning2014stanford}. 
We then filtered the datasets by excluding reports where (1) no findings or impression (i.e., summary) section can be found; (2) multiple findings or impression sections can be found but cannot be aligned; or (3) the findings have fewer than 10 words or the impression has fewer than 2 words. Lastly, we replaced all date and time mentions with special tokens (e.g., \textit{$<$DATE$>$}).

For both datasets, we stratified them over time into training, dev and test splits.
We employed this stratification strategy to test whether our model generalizes to future data when trained on historical data.
We show the time coverage of each split in \reftab{time-coverage}.

\begin{table}[t]
  \centering
  \setlength{\tabcolsep}{0.3em}
  \begin{tabular}{lccc}
    \toprule
    Variable & PG Baseline & \rlrc & $\Delta$ \\
    \midrule
    No Finding & 91.0 & 92.0 & \ph{0}+1.0$^*$ \\
    Cardiomegaly & 21.1 & 33.8 & +12.7$^*$ \\
    Airspace Opacity & 80.4 & 83.5 & \ph{0}+3.1$^*$ \\
    Edema & 73.4 & 80.2 & \ph{0}+6.8$^*$ \\
    Pneumonia & 63.5 & 69.2 & \ph{0}+5.7$^*$ \\
    Atelectasis& 60.5 & 66.5 & \ph{0}+6.0$^*$ \\
    Pneumothorax & 89.7 & 93.2 & \ph{0}+3.5$^*$ \\
    Pleural Effusion & 74.3 & 79.9 & \ph{0}+5.6$^*$ \\
    \midrule
    Macro Avg. & 69.3 & 74.8 & \ph{0}+5.5$^*$ \\
    \bottomrule
  \end{tabular}
  \caption{Test set performance for each variable on the RIH dataset. All numbers are \fone{} scores. $*$ marks statistically significant improvements with $p < .01$ under a bootstrap test.}
\label{tab:var-rih}
\end{table}

\section{Model Implementation and Training Details}
\label{sec:implementation}

For the baseline background-augmented pointer-generator model, we use its open implementation.%
\footnote{\url{https://github.com/yuhaozhang/summarize-radiology-findings}}
We use a 2-layer LSTM as the findings encoder, 1-layer LSTM as the background encoder, and a 1-layer LSTM as the decoder. 
For all LSTMs we use a hidden size of 200.
For the embedding layer we use 100-dimensional GloVe vectors~\cite{pennington2014glove} which we pretrained on about 4 million radiology reports.
We apply dropout~\cite{srivastava2014dropout} with $p=0.5$ to the embeddings.
At decoding time, we use the standard beam search with a beam size of 5 and a maximum decoding length of 50.

For the training and finetuning of the models, we use the Adam optimizer~\cite{kingma2014adam} with an initial learning rate of $1e^{-3}$.
We use a batch size of 64 and clip the gradient with a norm of 5.
During training we evaluate the model on the dev set every 500 steps and decay the learning rate by 0.5 whenever the validation score does not increase after 2500 steps.
Since we want the model outputs to have both high overlap with the human references and high factual correctness, for training we always use the average of the dev ROUGE score and the dev factual \fone{} score as the stopping criteria.
We tune the scalar weights in the loss function on the dev sets and use weights of $\lambda_1 = 0.97$, $\lambda_2 = 0.97$ and $\lambda_3 = 0.03$ for both datasets.

% qualitative examples
\begin{figure}[t!]
  \centering
	% \small
\scriptsize
\def\arraystretch{1.5}
\begin{tabular}{ | p{0.45\textwidth} | }
\hline
Stanford Dataset\\
\hline
% test example 12928
{\bf Background:}
radiographic examination of the chest: $<$date$>$ $<$time$>$. clinical history: $<$age$>$ years of age, with concern for pulmonary edema. procedure comments: 3 single views of the chest...
\vspace{0.5em}
\newline
{\bf Findings}:
in the first chest radiograph from $<$date$>$ at $<$time$>$ there is interval intubation. left arm-picc line remains in place. grossly unchanged persistent cardiomegaly, bilateral pleural effusion, and mild pulmonary edema. severe djd of the left gh joint is noted. in the second chest radiograph there is interval placement of a trialysis catheter in the left ij. no other significant changes are noted. in the third chest radiograph from $<$date$>$ at $<$time$>$ there is an increased left basilar opacity likely reflecting basilar consolidation, atelectasis or aspiration.
\\
\hline
{\bf Human}:
in the final chest radiograph there is \dblue{\ul{increased left basilar opacity likely reflecting basilar consolidation, atelectasis or aspiration}}.
\\
\hline
{\bf PG Baseline}: interval intubation with placement of a trialysis catheter in the left ij. grossly unchanged cardiomegaly, bilateral pleural effusion, and mild pulmonary edema.
\\
\hline
{\bf \rlrc{}}: interval placement of a trialysis catheter in the left ij. \dblue{\ul{an increased left basilar opacity likely reflecting basilar consolidation, atelectasis or aspiration or aspiration}}.
\\ 

\hhline{|=|}

RIH Dataset
\\
\hline
% test example 4553
{\bf Background:}
post op cardiac surgery - check lines and tubes. technique: single view of the chest obtained at $<$time$>$ $<$date$>$...
\vspace{0.5em}
\newline
{\bf Findings}:
lines/tubes: right ij sheath with central venous catheter tip overlying the svc. on initial radiograph, endotracheal tube between the clavicular heads, and enteric tube with side port at the ge junction and tip below the diaphragm off the field-of-view; these are removed on subsequent film. mediastinal drains and left thoracostomy tube are unchanged. lungs: low lung volumes. retrocardiac airspace disease, slightly increased on most recent film. pleura: small left pleural effusion. no pneumothorax. heart and mediastinum: postsurgical widening of the cardiomediastinal silhouette. aortic arch calcification. bones: intact median sternotomy wires.
\\
\hline
{\bf Human}: \dblue{\ul{left basilar airspace disease and small left pleural effusion}}. lines and tubes positioned as above.
\\
\hline
{\bf PG Baseline}: lines and tubes as above. retrocardiac airspace disease, which may \dorange{\uwave{represent atelectasis, aspiration, or pneumonia}}.
\\
\hline
{\bf \rlrc{}}: lines and tubes as described above. \dblue{\ul{retrocardiac airspace disease}}, slightly increased on most recent film. \dblue{\ul{small left pleural effusion}}.
\\ 
\hline

\end{tabular}
  \caption{More examples from the test splits of both datasets along with human, PG baseline and \rlrc{} summaries. 
  In the first example, the baseline output successfully copied content from the context, but missed \dblue{\ul{important observations}}.
  In the second example, the baseline output included some \dorange{\uwave{spurious facts}} that were not mentioned, and again neglected some \dblue{\ul{important observations}}.
  In neither examples the \rlrc{} outputs make perfect summaries, but they represent better summaries than the baseline outputs.}
  \label{fig:more-examples}
\end{figure}

For the extractive LexRank and BanditSum models, we use their open implementations.%
\footnote{\url{https://github.com/miso-belica/sumy}; \url{https://github.com/yuedongP/BanditSum}}
For the BanditSum extractive summarization model, we use default values for all hyperparameters as in \citet{dong2018banditsum}.
For both models we select the top 3 scored sentences to form the summary, which yields the highest ROUGE-L scores on the dev sets.

For ROUGE evaluation, we use the Python ROUGE implementation released by Google Research.%
\footnote{\url{https://github.com/google-research/google-research/tree/master/rouge}}
We empirically find it to provide very close results to the original Perl ROUGE implementation by~\citet{lin2004rouge}.

\section{Fine-grained Correctness Results on the RIH Dataset}
\label{sec:var-rih}

We show the fine-grained factual \fone{} scores for all variables on the RIH dataset in \reftab{var-rih}.

\section{More Examples with Baseline and System Generations}
\label{sec:more-examples}

In \reffig{more-examples} we present more examples from both datasets along with the generations from the baseline system and our approach.

\end{document}